\documentclass[letterpaper]{article} % DO NOT CHANGE THIS
\usepackage{aaai25}  % DO NOT CHANGE THIS
\usepackage{times}  % DO NOT CHANGE THIS
\usepackage{helvet}  % DO NOT CHANGE THIS
\usepackage{courier}  % DO NOT CHANGE THIS
\usepackage[hyphens]{url}  % DO NOT CHANGE THIS
\usepackage{graphicx} % DO NOT CHANGE THIS
\usepackage{natbib}  % DO NOT CHANGE THIS AND DO NOT ADD ANY OPTIONS TO IT
\usepackage{caption} % DO NOT CHANGE THIS AND DO NOT ADD ANY OPTIONS TO IT
\usepackage{algorithm}
\usepackage{algorithmic}

\usepackage{amsmath}
\usepackage[table]{xcolor}
\usepackage{xcolor}
\usepackage{amssymb}

\usepackage{listings}% 语法高亮和代码样式设置方面更加强大和灵活

\usepackage{newfloat}
\usepackage{listings}
\DeclareCaptionStyle{ruled}{labelfont=normalfont,labelsep=colon,strut=off} % DO NOT CHANGE THIS
\floatstyle{ruled}
\newfloat{listing}{tb}{lst}{}
\floatname{listing}{Listing}
\nocopyright

\title{Flatten: Video Action Recognition is an Image Classification task}
\author{
    Junlin Chen\textsuperscript{\rm 1}, 
    Chengcheng Xu\textsuperscript{\rm 1}\equalcontrib, 
    Yangfan Xu\textsuperscript{\rm 1}\equalcontrib, 
    Jian Yang\textsuperscript{\rm 1}\equalcontrib,  \\
    Jun Li\textsuperscript{\rm 1}\thanks{Corresponding Author.}, 
    Zhiping Shi\textsuperscript{\rm 1}\thanks{Corresponding Author.}
}
\affiliations{
    \textsuperscript{\rm 1}Information Engineering College, Capital Normal University\\
    No.56 West Third Ring Road North\\
    Haidian District, Beijing 100048, China\\
}

\usepackage{bibentry}
\begin{document}

\maketitle

\begin{abstract}

In recent years, video action recognition, as a fundamental task in the field of video understanding, has been deeply explored by numerous researchers. 
Most traditional video action recognition methods typically involve converting videos into three-dimensional data that encapsulates both spatial and temporal information, subsequently leveraging prevalent image understanding models to model and analyze these data.
However, these methods have significant drawbacks. Firstly, when delving into video action recognition tasks, image understanding models often need to be adapted accordingly in terms of model architecture and preprocessing for these spatiotemporal tasks;
Secondly, dealing with high-dimensional data often poses greater challenges and incurs higher time costs compared to its lower-dimensional counterparts. 
To bridge the gap between image-understanding and video-understanding tasks while simplifying the complexity of video comprehension, we introduce a novel video representation architecture, \textbf{Flatten}, which serves as a plug-and-play module that can be seamlessly integrated into any image-understanding network for efficient and effective 3D temporal data modeling. 
Specifically, by applying specific flattening operations (e.g., row-major transform), 3D spatiotemporal data is transformed into 2D spatial information, and then ordinary image understanding models are used to capture temporal dynamic and spatial semantic information, which in turn accomplishes effective and efficient video action recognition.
Extensive experiments on commonly used datasets (Kinetics-400, Something-Something v2, and HMDB-51) and three classical image classification models (Uniformer, SwinV2, and ResNet), have demonstrated that embedding Flatten provides a significant performance improvements over original model.
% the effectiveness and generality of our approach.
\end{abstract}


\section{Introduction}

In computer vision, for a long period, video modeling methods based on three-dimensional data have dominated the field of video action recognition. These methods can be broadly classified into three categories: a).Methods based on 3D convolutional backbone networks, such as C3D~\cite{tran2015learning}, I3D~\cite{carreira2017quo}, R(2+1)D~\cite{tran2018closer}, X3D~\cite{feichtenhofer2020x3d}, etc. b)Methods based on temporal fusion with 2D convolutions, such as Two-Stream Networks~\cite{Twostream}, TSN~\cite{wang2016temporal}, TDN~\cite{wang2021tdn}, SlowFast~\cite{feichtenhofer2019slowfast}, etc. c). Methods based on transformer backbone networks, such as TimeSformer~\cite{bertasius2021space}, Vivit~\cite{arnab2021vivit}, Video swin transformer~\cite{liu2022video}, etc.

\begin{figure}[!t]
    \centering
    \includegraphics[width=\columnwidth]{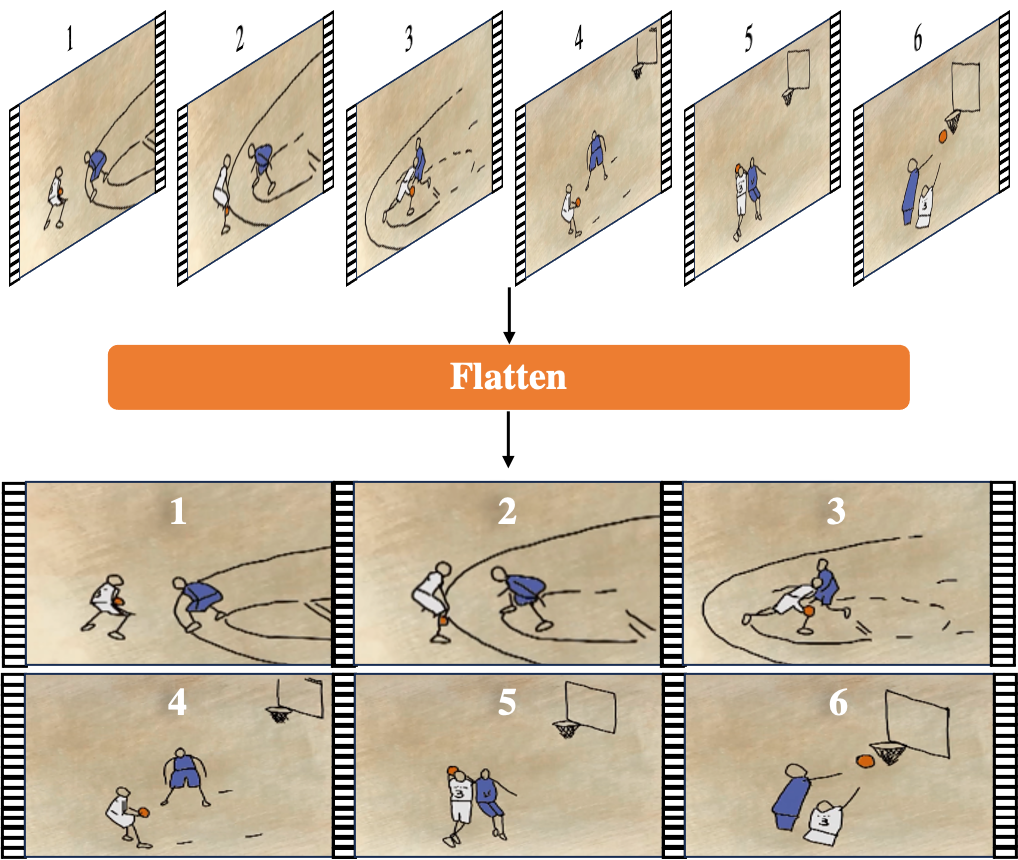}
    \caption{The Flatten operation converts image sequences into single-frame two-dimensional images, enabling the processing of video-understanding tasks like reading a comic strip.}
    \label{fig:intro}
\end{figure}

Methods based on 3D convolutional backbone networks utilize 3D convolutional kernels to model the dependencies between adjacent spatial and temporal regions. By stacking a sufficient number of network layers, they achieve global dependency modeling of video information. Methods based on temporal fusion with 2D convolutions decompose video tasks into short-term and long-term temporal modeling, and then fuse the short-term and long-term information to achieve comprehensive modeling of video information.
Thanks to the transformer~\cite{vaswani2017attention} architecture's superior long-sequence modeling capabilities, sparking significant interest among researchers in exploring the use of transformers for spatiotemporal modeling in videos. They have investigated various approaches~\cite{tong2022videomae,wang2024internvideo2} to balance the effectiveness and efficiency of video spatiotemporal modeling, utilized additional data for joint training to enhance modeling performance, and constructed large models for video-understanding.

Although these three types of methods employ different concepts for video sequence modeling, they also share some common aspects. 
Firstly, these methods treat video action recognition as a three-dimensional data classification problem, attempting to infer video actions from the spatiotemporal relationships conveyed by the video. 
Secondly, the original Neural network model and pre-trained models for these methods are derived from image classification tasks. They adapt data preprocessing methods and model structures to fit the characteristics of video action recognition tasks, enabling image models to be used for video-understanding tasks.
These characteristics also reflect certain drawbacks of the aforementioned methods. Since the original Neural network models are designed for two-dimensional image modeling, they typically require corresponding modifications for three-dimensional video modeling. 
This adds to the workload and, to some extent, separates image tasks from video tasks. Additionally, modeling relationships between high-dimensional data is more challenging compared to modeling relationships between low-dimensional data.

A better understanding of the intrinsic logic of action recognition is the premise for effectively solving the problem of video action recognition, i.e., how to distinguish one action from another through video. From an empirical perspective, two factors typically influence action recognition.
One factor is the static scene information in the frame. For instance, different behaviors of individuals in a car can be distinguished based on whether a person is holding the steering wheel. The behavior of a person holding the steering wheel is defined as driving, while the behavior of a person not holding the steering wheel is defined as riding.
Another factor is the relative motion or position change in the frames, essentially a dynamic relative change. For example, determining whether a person on a playground is walking or running can be done by observing the amplitude and speed of their body movements, reflecting relative motion and position changes.
For the first scenario, to achieve video action recognition, modeling the static images of the video is sufficient. For example, in UCF101~\cite{soomro2012ucf101}, action recognition accuracy can exceed 80\% by solely recognizing images.
For the second scenario, past practices in tasks such as object detection~\cite{carion2020end,wang2024yolov10} and semantic segmentation~\cite{cao2022swin} have proven that deep learning models can recognize different objects in different spatial locations through various features. This also suggests a possibility that deep learning models can differentiate between the actions of the same object in images and establish relationships between different actions and between the subject and the environment. This phenomenon is akin to how humans understand actions in a comic; the model can comprehend action information through images, as shown in figure \ref{fig:intro}.

Based on the aforementioned review of past research and the analysis of the video action recognition task, we propose \textbf{Flatten}. \textbf{Flatten} transforms the video action recognition task into an image classification task, which to some extent, blurs the boundary between image-understanding tasks and video-understanding tasks, thereby reducing the complexity of video action recognition. In summary, our contributions are as follows:

\begin{itemize}
    % \item We propose the Flatten method, which for the first time, transforms the video action recognition task into an image classification task. By converting the video action recognition task of modeling three-dimensional data relationships into an image classification task of modeling two-dimensional data relationships, we blur the boundary between image-understanding tasks and video-understanding tasks, offering new possibilities for solving video-understanding tasks.
    \item To the best of our knowledge, we are the first to propose the Flatten method, a plug-and-play parameter-free transform that transforms spatiotemporal modeling into spatial modeling, blurs the boundaries between image and video understanding tasks, and provides a novel and efficient solution to address video action recognition. 
    \item We construct three different instances of Flatten and demonstrate through comparative experiments on these three instances that image-understanding models can learn data mapping relationships to achieve the function of modeling temporal information in the spatial dimension.
    \item Extensive experiments conducted on classical CNN, Transformer, and hybrid CNN-Transformer models for image classification have consistently demonstrated that incorporating the Flatten transformation leads to significant performance gains. In particular, ``Uniformer(2D)-S + Flatten'' achieves the best recognition performance of 81.1 on the Kinetics400 dataset when compared to other state-of-the-art methods with the same amount of computation.
\end{itemize}

\section{Related Work}

\begin{figure*}[t]
\centering
\includegraphics[width=0.9\linewidth]{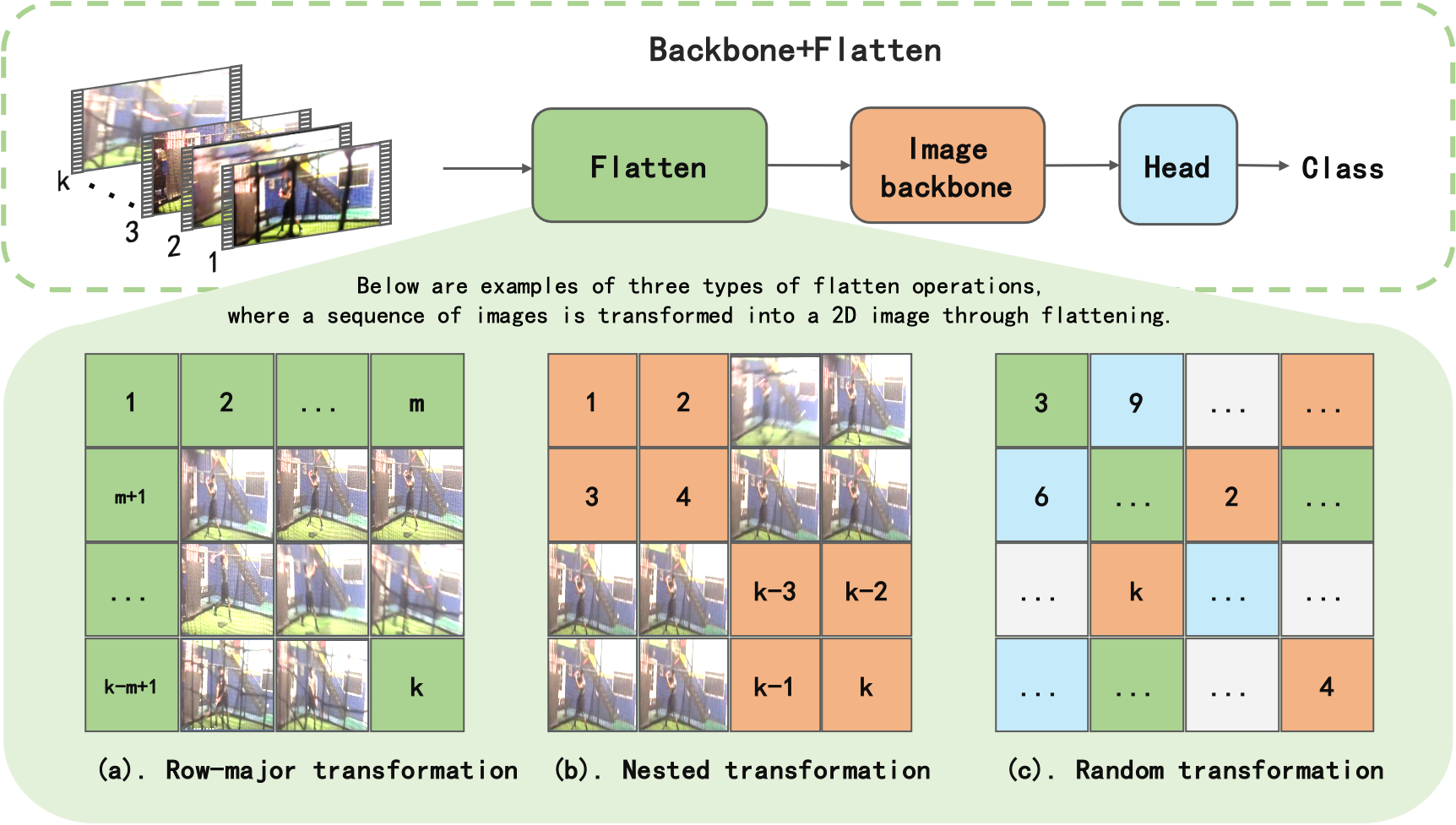} % Reduce the figure size so that it is slightly narrower than the column. Don't use precise values for figure width.This setup will avoid overfull boxes.
\caption{An overview of action recognition network embedding Flatten operation, which converts a sequence of images into a single-frame image, enabling image-understanding models to be easily applied to video-understanding tasks. Meanwhile, we provide three examples of Flatten operations: (a) row-major transformation, (b) nested transformation, and (c) random transformation.}
\label{pipline}
\end{figure*}

\subsection{Video Understanding}

Temporal action information is particularly important in the field of video-understanding. Numerous works revolve around capturing spatiotemporal information in videos, with researchers attempting to model this information to establish robust video-understanding capabilities. The Two-Stream Network~\cite{Twostream} and its variants~\cite{feichtenhofer2016convolutional} enhance the network's video-understanding ability by utilizing optical flow information between adjacent frames to enrich spatiotemporal information. However, a consequent issue is that obtaining optical flow information is very costly.
Some works use 3D convolutions to model video information, capturing spatiotemporal information by stacking network depth. This approach started with C3D~\cite{tran2015learning}, and subsequent researchers proposed methods like I3D~\cite{carreira2017quo} and R2+1D~\cite{tran2018closer} to achieve faster convergence and lower computational complexity for 3D convolution-based methods.
With the excellent capabilities of Transformers~\cite{vaswani2017attention} in temporal modeling, multimodal fusion, and large-scale models, numerous works have begun exploring the application of Transformers in the video domain. Methods such as TimeSformer~\cite{bertasius2021space}, Vivit~\cite{arnab2021vivit}, Video swin transformer~\cite{liu2022video}, Video MAE~\cite{tong2022videomae}, and InternVideo~\cite{wang2024internvideo2} have emerged.
Despite the differences among the aforementioned methods, they also share some common characteristics. Firstly, video-understanding models rely on fundamental image models. Secondly, they treat video tasks as three-dimensional information modeling tasks, distinct from image tasks. Finally, to apply image models to video tasks, additional model adjustments or data processing are usually required.

\subsection{Image Understanding}

As a fundamental task in computer vision, image-understanding has consistently attracted the attention of numerous researchers. Particularly after the introduction of AlexNet, which employed convolutional neural networks (CNNs), deep learning methods have garnered significant interest from researchers. Subsequently, many works have explored the organization of convolutional neural networks. ResNet~\cite{he2016deep}, with its residual connections and residual blocks, established a classical convolutional neural network, leading to extensive subsequent research into the impact of network depth and width on model performance.
Given the exceptional performance of Transformers~\cite{vaswani2017attention} in natural language processing, some works have applied Transformers to image-understanding, enabling the model's receptive field to be independent of network depth. Recent works~\cite{dai2021coatnet,li2023uniformer} have attempted to combine the strengths of convolutional neural networks and Transformers, leveraging the inductive biases of CNNs and the global modeling capabilities of Transformers to enhance model performance.

\section{Method}

In this section, we first present the definition of the Flatten method, followed by providing several examples of different types of Flatten.

\subsection{Formulation}

As described above, the starting point of Flatten is to transform the video action recognition task from a three-dimensional data modeling task into a two-dimensional data modeling task, thereby blurring the boundary between image-understanding and video-understanding. Following the starting point of Flatten, we define a general Flatten operation in deep neural networks as follows:

\begin{equation}
\label{eq:1}
    Y = g(f(x,y,z)).
\end{equation}

In the equation \ref{eq:1}, $g(x,y)$ represents a general image-understanding neural network, and $f(x,y,z)$ denotes a transformation rule. 
Where $x,y$, and $z$ represent the width, height, and number of frames of a sequence of images, respectively.
Through this transformation, the video data represented by a set of images $ I \in \mathbb{R}^{x \times y \times z} $ can be converted into video data represented by a single image $ I' \in \mathbb{R}^{x' \times y'} $.
In this description, we temporarily disregard the channel information $C$ of the images and focus more on the spatial representation of temporal information $Z$ in the video data, i.e., $(x,y,z)\rightarrow(x',y')$.
We aim to construct a mapping between three-dimensional and two-dimensional data through $f(x,y,z)$,  enabling the deep learning model to learn this mapping relationship based on image-understanding, thereby achieving temporal modeling in the spatial dimension.

The work most similar to the Flatten method is the "Uniform frame sampling" method in Vivit~\cite{arnab2021vivit}. This method encodes each frame in a set of images containing video information into tokens using the VIT~\cite{dosovitskiy2020vit} method and then concatenates all the tokens from the images to represent the entire video segment. After this, an image-understanding model like VIT can be used for video-understanding. However, this method has two problems: first, its applicability is limited and can only be used with specific neural networks, such as Transformer-like networks. Second, due to the large amount of redundant information present in videos and the limited range of models this method can be applied to, it is very costly to use this approach.

Since the Flatten method does not alter the model structure, it is very flexible and can be easily applied to various image-understanding models, including but not limited to ResNet, Swin-Transformer, and Uniformer. Relying on a significant amount of image-understanding work(Swin, Uniformer), it does not face the exponential growth in computational complexity that the "Uniform frame sampling" method in Vivit encounters. Therefore, we can use the Flatten operation to easily construct an image-understanding model to comprehend temporal information in videos, thereby achieving accurate video action recognition.

\subsection{Instantiations}

Next, we describe several types of $f(x,y,z)$. Interestingly, our experimental results show that the model is not sensitive to several fixed transformation rules of $f(x,y,z)$, but it is sensitive to unfixed transformation rules of $f(x,y,z)$.
This further indicates that the model learns this mapping relationship $f(x,y,z)$, thereby achieving temporal information modeling in the spatial dimension.

\textbf{Row-major transformation.} The most empirically intuitive data transformation method is row-major ordering. As shown in Figure \ref{pipline} (a). This transformation method is akin to converting a video into a comic strip, i.e., unfolding the data along the row direction through a transformation function $f$.
In this paper, we define $f$ as follows:

\begin{align}
    f(x_i,y_j,z_k) &= \left\{\begin{matrix}
     x'_i = x_i + l_k \times  w \\
     y'_j = y_j + c_k \times  h
\end{matrix}\right., \label{eq:2} \\
l_k &= \left\lfloor z_k\ /\ m\right\rfloor, \label{eq:3} \\
c_k &= z_k \%~ m. \label{eq:4}
\end{align}

Here, $(x_i, y_j, z_k)$ represents the video information represented by a set of images before the transformation, and $(x', y')$ represents the video information represented by a single image after the transformation.
In Equation \ref{eq:3} and Equation \ref{eq:4}, $m$ denotes that the width and height of the transformed image $I' \in \mathbb{R}^{x' \times y'} $ are composed of m frames of the original image $i \in \mathbb{R}^{x \times y}$, and $m \times m$ equals the total number of frames in this set of images $I \in \mathbb{R}^{x \times y \times z}$.
This transformation is illustrated in Figure \ref{pipline} (a).

\textbf{Nested transformation.} Inspired by methods such as SlowFast~\cite{fan2020pyslowfast} and TDN~\cite{wang2021tdn}, we attempt to consider the balance between short-term and long-term temporal relationships during the data transformation process, leading to the development of the second data transformation method. In this paper, it is represented by Equation 9:

\begin{align}
    I &= \left\{{i}_{1},{i}_{2},{i}_{3},...{i}_{k}\right\}, \label{eq:5} \\
    {q} &= {K}/{N}, \label{eq:6} \\
    {i^\prime}_n &= f\left(I_n\right), \label{eq:7} \\
    {I'} &= \left\{{{i'}}_{1},{{i'}}_{2},{{i'}}_{3},...{{i'}}_{n}\right\}, \label{eq:8} \\ 
    {I''} &= {f}({I'}), \label{eq:9}
\end{align}
\begin{equation}
    \label{eq:10}
    I_n=\left\{i_{\left(n-1\right) \times q+1},i_{\left(n-1\right)\times q+2}, i_{\left(n-1\right)\times q+3},\ldots i_{n\times q}\right\}.
\end{equation}

Among them, $I$ represents an ordered image sequence containing $K$ images that encapsulate the video information. By partitioning, $I$ is divided into $N$ sub-image sequences, with the $n$-th sub-sequence represented as $I_n$, and each sub-sequence contains $q$ images. Each sub-sequence $I_n$ undergoes a row-major transformation $f$, resulting in an ordered image sequence $I'$ represented by $N$ images, with each image $i'_n$ transformed from a sub-image sequence $I_n$. Finally, by applying another row-major transformation to $I'$, we obtain the nested transformation image $I''$. An example of this transformation is illustrated in Figure \ref{pipline} (b).

\begin{table*}[t]
    \centering
    % \resizebox{\textwidth}{!}{}
    \begin{tabular}{l | l | l | c | c | c}
    \hline
        Method & Pretrain & \#Frame & GFLOPs & Top-1 & Top-5 \\
    \hline
       I3D R101 + NL~\cite{carreira2017quo} & IN-1K & 224$\times$128$\times$3$\times$10 & 359$\times$30 & 77.7 & 93.3 \\
        SlowFast R101~\cite{feichtenhofer2019slowfast} & - & 224$\times$(16+64)$\times$3$\times$10 & 213$\times$30 & 78.9 & 93.5 \\
        SlowFast R101 + NL~\cite{feichtenhofer2019slowfast}& - & 224$\times$(16+64)$\times$3$\times$10 & 234$\times$30 & 79.8 & 93.9 \\
    \rowcolor{gray!20}
        \textbf{ResNet101 + Flatten} & IN-1K & \textbf{224$\times$16$\times$3$\times$10} & 125$\times$30 & 76.2 & 92.9 \\
    \hline
    \hline
       ViT-B-VTN~\cite{neimark2021video}  & IN-21K & 224$\times$250$\times$1$\times$1 & 3992 & 78.6 & 93.7 \\
      TimeSformer-L~\cite{bertasius2021space} & IN-21K & 224$\times$96$\times$3$\times$1 & 2380$\times$3 & 80.7 & 94.7 \\
      ViViT-L~\cite{arnab2021vivit} & IN-21K & 224$\times$16$\times$3$\times$4 & 1446$\times$12 & 80.6 & 94.7 \\
       Video Swin-T~\cite{liu2022video} & IN-1K & 224$\times$32$\times$3$\times$4 & 88$\times$12 & 78.8 & 93.6 \\
    \rowcolor{gray!20}
       \textbf{SwinV2-T + Flatten} & IN-1K & \textbf{192$\times$16$\times$1$\times$4} & 70.9$\times$4 & 78.6 & 93.9 \\
    \hline
    \hline
        Uniformer(3D)-XXS~\cite{li2023uniformer}  & IN-1K & 160$\times$8$\times$1$\times$1 & 3.3 & 71.4 & 89.4 \\
       Uniformer(3D)-XXS~\cite{li2023uniformer}  & IN-1K & 160$\times$16$\times$1$\times$1 & 6.9 & 75.1 & 91.5 \\
       Uniformer(3D)-S~\cite{li2023uniformer}  & IN-1K & 224$\times$8$\times$1$\times$4 & 17.6$\times$4 & 78.4 & 93.3 \\
       Uniformer(3D)-S~\cite{li2023uniformer}  & IN-1K & 224$\times$16$\times$1$\times$4 & 41.8$\times$4 & 80.8 & 94.7 \\
    \rowcolor{gray!20}
        \textbf{Uniformer(2D)-XXS + Flatten}& IN-1K & 160$\times$9$\times$1$\times$1 & 10.5 & 72.9 & 90.9 \\
    \rowcolor{gray!20}
        \textbf{Uniformer(2D)-XXS + Flatten} & IN-1K & 160$\times$16$\times$1$\times$1 & 22.9 & 75.9 & 92.3 \\
    \rowcolor{gray!20}
        \textbf{Uniformer(2D)-S + Flatten} & IN-1K & 224$\times$9$\times$1$\times$4 & 47.5$\times$4 & 79.7 & 94.1 \\
    \rowcolor{gray!20}
        \textbf{Uniformer(2D)-S + Flatten} & IN-1K & 224$\times$16$\times$1$\times$4 & 107.3$\times$4 & \textbf{81.1} & \textbf{95.3} \\
        
    \hline
    \end{tabular}
    \caption{Comparison with state-of-the-art on the Kinetics400 dataset. \#Frame AxBxCxD represents A resolution, B frames, C spatial crops, and D temporal crops. '-' indicates unknown data.}
    \label{tab:K400}
\end{table*}

\textbf{Random transformation.} In real life, sometimes the sequence of ordered objects can be disturbed without affecting the understanding of the original meaning. For example, "AAAI is a renowned academic conference" can be rearranged as "a renowned AAAI is a conference academic" without significantly affecting people's understanding of the sentence. With a certain background knowledge, we can correct the errors in the sentence. Inspired by this, we perform a random transformation on the ordered image sequence. On the one hand, we aim to compare the random transformation with the two aforementioned fixed rule transformations to investigate whether the model can learn the transformation rules to model temporal relationships in the spatial dimension. On the other hand, we attempt to understand how the model's ability to understand video is affected when the image sequence is disturbed, and whether it possesses a certain "error correction capability." The transformation rule is shown in the following equation:

\begin{align}
    I' &= random(I), \label{eq:11} \\
    {I}^\prime &= \left\{{i}_{3},{i}_{k},{i}_{8},...{i}_{2}\right\}, \label{eq:12} \\
    {I}^{\prime\prime} &= {f}\left({I}^\prime\right). \label{eq:13}
\end{align}

Here, we first shuffle the temporal sequence of the ordered image sequence $I$, obtaining a disordered set of images $I'$. Finally, applying a row-major transformation results in the randomly transformed output $I''$. We provide an illustration of this transformation, as shown in Figure \ref{pipline} (c).

% \section{Empirical Evaluation}
\section{Experiments}

In this section, we introduce the validation experiments and details and analyze them. The experimental section will be presented in the following order: \textbf{Experimental Setup}, \textbf{Ablation Experiments on the Flatten}, \textbf{Comparison to state-of-the-art} and \textbf{Visualization}.

% \subsection{Experimental Setup}
\subsection{Experimental Setup}
\label{Experimental Setup}

Our experiments involve three mainstream network models, three types of Flatten operations, and three mainstream video action recognition datasets. The three network models are ResNet~\cite{he2016deep}, Swinv2~\cite{liu2022swin}, and Uniformer~\cite{li2023uniformer}, representing convolutional neural networks, Transformers, and a fusion of CNN and Transformer networks, respectively. We apply the Flatten operation to different neural networks to demonstrate the method's generality. The three types of Flatten operations are the three transformations introduced in the methodology section, which can be classified into fixed rule transformations and random transformations. Ablation experiments with the three Flatten operations demonstrate the network models' ability to learn spatiotemporal mapping relationships and show that the models can model temporal relationships through spatial information. The three mainstream video action recognition datasets are Kinetics400~\cite{kay2017kinetics}, Something-Something V2~\cite{goyal2017something}, and HMDB51~\cite{kuehne2011hmdb}. By showcasing performance on these three datasets, we prove the effectiveness of the Flatten method.

\textbf{Network Models:} First, we briefly introduce the network models used in the experiments and their training parameters.

\begin{table}
    \centering
    \begin{tabular}{c | c | c| c}
        \hline    
        Flatten variants & Top-1 & Top-5 & \#Frame \\
        \hline
        Row-major & 75.87 & 92.3 & 160$\times$16$\times$1$\times$1 \\
        Nested & 75.83 & 92.6 & 160$\times$16$\times$1$\times$1 \\
        Random & 71.1 & 90.0 & 160$\times$16$\times$1$\times$1 \\
        \hline
        Uniformer(3D)-XXS & 75.1 & 91.5 & 160$\times$16$\times$1$\times$1 \\
        \hline
    \end{tabular}
    \caption{The impact of different Flatten variants on the video-understanding capability of the Uniformer-XXS model. We report the Top-1 and Top-5 accuracies of different variants on the Kinetics400 dataset, as well as the accuracy of the original Uniformer-XXS model.}
    \label{tab:ablation}
\end{table}

ResNet~\cite{he2016deep} is a classic convolutional neural network widely used in deep learning. It utilizes residual connections and blocks to mitigate degradation in deep neural networks, enabling deeper models. In our experiments, we use ResNet101 as the backbone.

Swin-TransformerV2~\cite{liu2022swin} is a successful application of Transformer in the field of image-understanding. By utilizing a sliding window mechanism, it decomposes global attention into local window attention and window information merging, addressing the problem of exponential growth in computational cost with image size seen in the ViT method. This allows the model to better train on high-resolution images. In our experiments, we use SwinV2-T as the backbone.

Uniformer~\cite{li2023uniformer} combines convolutional neural networks and Transformers. It leverages convolutional networks' inductive bias and Transformers' global modeling, applying convolutions in shallow layers and Transformers in deep layers. This hybrid approach excels in computer vision tasks. We conduct experiments on Uniformer-XXS and Uniformer-S in this paper. In the paper, the original Uniformer model used for video understanding is referred to as Uniformer (3D), while the Uniformer model used for image understanding is referred to as Uniformer (2D).

All our experimental code is based on modifications to the SlowFast code framework publicly available on GitHub from Facebook (Meta)~\cite{fan2020pyslowfast}. The specific experimental parameters for the three models will be provided in the appendix.

\begin{table}[t]
    \centering
    \begin{tabular}{l|l|c|c}
    \hline
        Method & Pretrain & Top-1 & Top-5 \\
    \hline
        TSN~\cite{wang2016temporal} & IN-1K & 30.0 & 60.5 \\
        SlowFast & - & 61.7 & - \\
        TDN & IN-1k & 65.3 & 89.5 \\
    \hline
        TimeSformer-HR & IN-21K & 62.5 & - \\
        ViViT-L & K400 & 65.4 & 89.8 \\
    \textcolor{black!50}{Video Swin-B} & \textcolor{black!50}{K400} & \textcolor{black!50}{69.6} & \textcolor{black!50}{92.7} \\
    \hline
    \hline
        UniFormer(3D)-S & K400 & 67.7 & 91.4 \\
    \rowcolor{gray!20}
        \textbf{UniFormer(2D)-S+Flatten} & K400 & 67.9 & 91.9 \\
    \hline
    
    \end{tabular}
    \caption{Comparison with State-of-the-Art on the Something-Something V2 Dataset}
    \label{tab:SSV2}
\end{table}

\textbf{Datasets:} We evaluate the performance of our proposed method on the following video classification datasets.

Kinetics400~\cite{kay2017kinetics} is a large-scale video action recognition dataset by DeepMind, containing about 260,000 ten-second video clips across 400 action categories. Sourced from YouTube and manually annotated, it is widely used for training and evaluating video action recognition models, serving as a key benchmark in video-understanding and computer vision research.

Something-Something V2 (SSV2)~\cite{goyal2017something} is a large video dataset for action recognition tasks, with 220,847 clips across 174 action categories. It focuses on interactions between objects, emphasizing fine-grained motion recognition. SSV2 is extensively used for training and evaluating video-understanding models, advancing temporal video analysis.

HMDB51~\cite{kuehne2011hmdb} is a benchmark dataset for human action recognition research by Harvard University, with 6,766 video clips across 51 action categories. Videos are sourced from movies, YouTube, and Google Videos, covering various everyday actions like "drinking," "running," and "dancing." Each category has about 100-200 clips, making HMDB51 a crucial dataset for training and evaluating video action recognition models.

\subsection{Ablation Study}
\textbf{Flatten Variants:} As introduced in Section Method, our experiments design three types of Flatten operations. These can be categorized into two groups. The first group consists of rule-based Flatten transformations, such as row-major transformation and nested transformation. Given a set of images $I \in \mathbb{R}^{x \times y \times z}$, each transformation will consistently yield the same $I' \in \mathbb{R}^{x' \times y'}$. The second group consists of non-rule-based Flatten transformations, such as random transformation, where the transformed data is random each time, with only a small probability that two transformations will be identical.

We chose to compare the effects of three Flatten operations on the model's ability to model temporal relationships in the spatial dimension using the Uniformer-XXS model on the Kinetics400 dataset. The objectives of this set of experiments are threefold: 1. To explore the impact of different Flatten transformations on the model's ability to model temporal relationships in the spatial dimension, especially the effects of different rule-based Flatten transformations on the model. 2. To investigate the impact of unordered image sequences on the model's video-understanding capability, as described in Section Method "Random Transformation". We aim to understand the model's "error-correction ability" for image sequences. 3. By comparing rule-based Flatten transformations with non-rule-based Flatten transformations, we attempt to demonstrate that the network can learn rule-based Flatten transformations $f$, thereby achieving temporal relationship modeling in the spatial dimension.

As shown in Table~\ref{tab:ablation}, both the row-major transformation and the nested transformation outperform the original Uniformer(3D)-XXS in terms of Top-1 accuracy, demonstrating the effectiveness of the Flatten method. Additionally, the accuracy difference between the row-major transformation(75.87\%) and the nested transformation(75.83\%) in rule-based Flatten transformations is minimal, with only a Top-1 accuracy difference of \textbf{0.04\%}. In contrast, the non-rule-based random transformation(70.9\%) has a Top-1 accuracy difference of \textbf{4.8\%} compared to the row-major transformation. This phenomenon indicates that deep learning models can achieve temporal relationship modeling in the spatial dimension by learning fixed Flatten transformation rules and are not sensitive to different rule-based Flatten transformations. Additionally, the model using random transformation is still able to converge, demonstrating that it possesses a certain degree of "error-correction ability".

\begin{figure*}[t]
    \centering
    \includegraphics[width=0.9\linewidth]{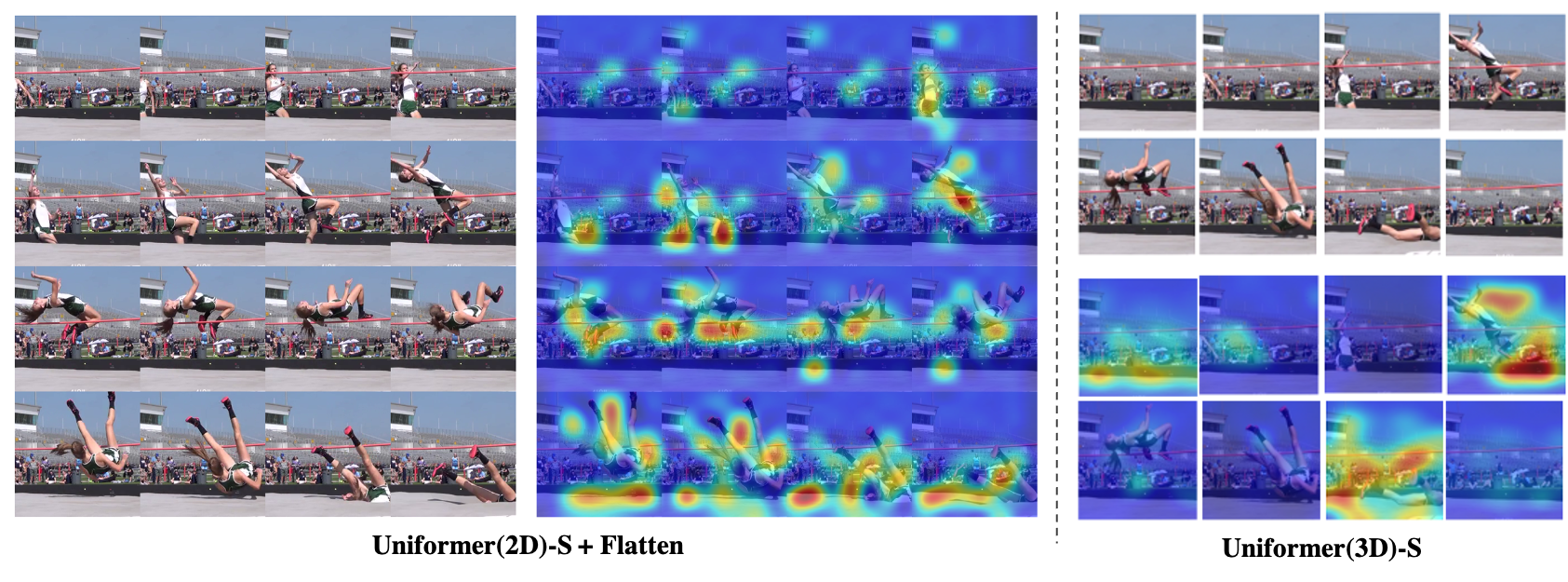}
    \caption{Heatmap visualization comparison. Figure~\ref{fig:heatmap} shows how the Uniformer(2D)-S+Flatten model performs temporal modeling in space. From the figure, we can clearly see that the Uniformer(2D)-S+Flatten model has a more sensitive observation of action information and stronger attention compared to the original Uniformer(3D)-S.}
    \label{fig:heatmap}
\end{figure*}

\begin{table}[t]
    \centering
    \begin{tabular}{l|l|c|c}
    \hline
        Method & Pretrain & Top-1 & Top-5 \\
    \hline
        Two-stream & IN-1K & 59.4 & - \\
        TSN~\cite{wang2016temporal} & IN-1K & 68.5 & - \\
        I3D & K400 & 74.8 & - \\
        MSNet~\cite{liu2020ms} & K400 & 77.4 & - \\
        VidTr-L~\cite{zhang2021vidtr} & K400 & 74.4 & - \\
    \hline
    \hline
        Uniformer(3D)-S & K400 & 77.5 & 96.1 \\
    \rowcolor{gray!20}
        \textbf{Uniformer(2D)-S+Flatten} & K400 & 82.9 & 96.9 \\
    \hline
    \end{tabular}
    \caption{Comparison with State-of-the-Art on the HMDB51 Dataset}
    \label{tab:HMDB51}
\end{table}

\subsection{Comparison with State-of-the-Art methods}

Based on the results of the ablation experiments in the previous section, we selected the row-major transformation as the Flatten operation for subsequent experiments. As described in Subsection Experimental Setup, to demonstrate the effectiveness and generality of the Flatten method, we will test its performance on three mainstream video action datasets (Kinetics400, SSV2, HMDB51) and three mainstream network models (ResNet, SwinV2, Uniformer).

Table~\ref{tab:K400} shows the performance of three different image-understanding network models using the Flatten method for video action recognition on the Kinetics400 dataset. In the XXS and S model sizes of Uniformer, the accuracy \textbf{81.1\%} of the Uniformer models using the Flatten method exceeds that of the 3D version of Uniformer(3D) used for video-understanding. Under conditions of smaller image resolution, fewer sampled frames, and lower computational cost, SwinV2-T achieved accuracy \textbf{78.6\%} comparable to that of the Video Swin-T. In convolutional neural networks, by directly feeding a sequence of images representing video information, transformed by the Flatten method, into an original ResNet101 network, we achieved an accuracy of \textbf{76.2\%}.

To further demonstrate the effectiveness and versatility of the Flatten method, we also conducted tests on the Something-Something V2 (SSV2) and HMDB51 datasets. As shown in Table~\ref{tab:SSV2} and Table~\ref{tab:HMDB51}, we evaluated the performance of Uniformer-S+Flatten on the SSV2 and HMDB51 datasets, respectively. These evaluations were conducted using pretrained models from the Kinetics400 dataset. The experimental results indicate that the Uniformer(2D)-S+Flatten method achieves superior Top-1 accuracy compared to the original Uniformer(3D)-S method. Specifically, Uniformer(2D)-S+Flatten achieved Top-1 accuracy of \textbf{67.9\%} on SSV2 and \textbf{82.9\%} on HMDB51. Comprehensive experiments demonstrate that the Flatten method is an effective and versatile data transformation method, enabling image-understanding models to be easily applied to video-understanding tasks without altering the fundamental model structure, while achieving competitive accuracy performance.

\begin{figure}[t]
    \centering
    \includegraphics[width=\linewidth]{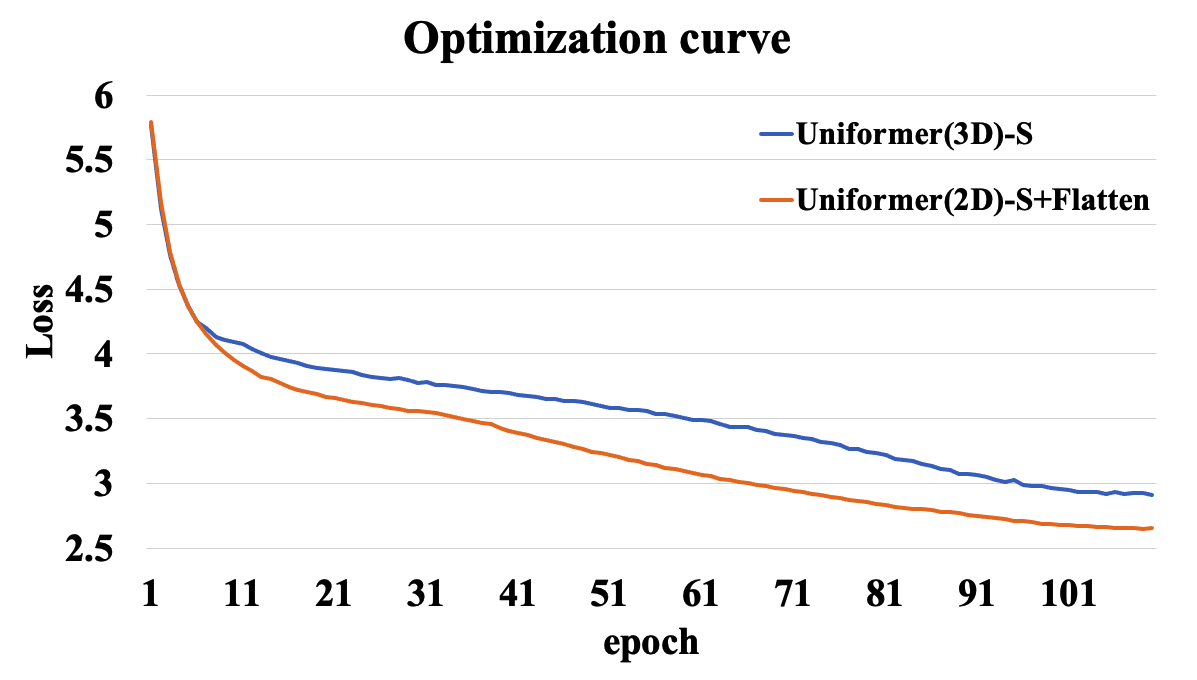}
    \caption{ Optimization curves of Uniformer(3D)-S and Uniformer(2D)-S+Flatten trained on the Kinetics400 dataset. Compared to Uniformer(3D)-S, Uniformer(2D)-S+Flatten demonstrates faster convergence and lower loss values under the same number of training epochs.}
    \label{fig:loss_curve}
\end{figure}

\subsection{Visualization}

To further validate the effectiveness of the Flatten method, we conducted several visualizations. In Figure~\ref{fig:heatmap}, we applied Grad-CAM~\cite{selvaraju2017grad} to generate heatmaps, analyzing how the model learns temporal relationships through spatial modeling. Figure~\ref{fig:loss_curve} shows the loss curves of Uniformer(3D)-S and Uniformer(2D)+Flatten trained on the Kinetics400 dataset, illustrating the effectiveness of the Flatten method. In the attention heatmaps, we can clearly see that the focus points strongly correlate with the actions displayed in the videos. The loss curves indicate that the model using the Flatten method achieves a lower loss and converges faster. In the final epoch, the loss value of Uniformer(2D)-S+Flatten is 9\% lower than that of Uniformer(3D)-S. These visualizations further corroborate the effectiveness of the Flatten method. More visualization results will be presented in the appendix.

\section{Conclusion}

We propose a data transformation method called Flatten, which for the first time treats video-understanding tasks as image-understanding tasks, thereby blurring the boundary between these two domains. Experiments on multiple mainstream video-understanding datasets and image-understanding neural networks have demonstrated the effectiveness and versatility of the Flatten method. In the future, we plan to further investigate the relationship between video and image data. Building on Flatten, we aim to explore joint training methods for images and videos to mitigate the issue of limited video data. Additionally, we intend to delve deeper into the "error correction ability" of models. As shown in Subsection "Ablation Study", the model can still accurately recognize some sequences even after the images are shuffled, which could lead to new methods for video data augmentation. Lastly, we hope to extend the Flatten method to other types of sequential data, broadening its range of applications.

\section{Appendix}

In the appendix, we will provide more detailed implementation specifics and supporting materials. These include the experimental platform, hyperparameter settings, visualization results, and key portions of the code.

\subsection{Implementation Details}

In our experiments, we built all training code using the PyTorch deep learning framework, with the training framework sourced from Facebook (Meta)'s open-source SlowFast~\cite{fan2020pyslowfast} project on GitHub. We modified this codebase to incorporate the Flatten method and conducted experiments across multiple datasets(K400~\cite{kay2017kinetics}, SSV2~\cite{goyal2017something}, HMDB51~\cite{kuehne2011hmdb}) and deep learning models(ResNet~\cite{he2016deep}, Swinv2~\cite{liu2022swin}, Uniformer). The training environment included an NVIDIA GTX 3090 GPU, with Python version 3.12.4.

Our experiments can be divided into three parts: (1) evaluating Uniformer with different model sizes on Kinetics-400, (2) testing Uniformer across various datasets(K400, SSV2, HMDB51), and (3) assessing different network models on the same dataset, Kinetics-400. The experimental parameters are detailed in Tables \ref{tab:1}, \ref{tab:2}, and \ref{tab:3}.

For all models trained on the K400 dataset, we use the pre-trained weights from the original models trained on the ImageNet1k dataset as initialization. For instance, we initialize the ResNet+Flatten model with the pre-trained weights obtained from the ImageNet1k dataset and then train it on the K400 dataset. For models trained on SSV2 and HMDB51, their pre-trained weights come from the training results of those models on the K400 dataset. For example, the initialization weights for Uniformer-S+Flatten come from the training results of Uniformer-S on the K400 dataset.

\subsection{Code}

In this section, we will present the key code for the Flatten method. The code is relatively short and does not modify the network model in any way. Therefore, it can be considered a simple, effective, and versatile method. By converting a sequence of images that represent both temporal and spatial information of a video into a single 2D image, the Flatten method reduces the boundary between video understanding tasks and image understanding tasks.

\begin{lstlisting}
def flatten(frames, flatten_type):
    assert frames.ndim == 4, 
    "Input tensor should have 4 dimensions."
    t = frames.shape[0]
    c = frames.shape[1]
    size = frames.shape[2]
    
    for n in range(1, length // 2 + 1):  
        m = length // n 
        if n*m == t and n==m:  
            break

    #Rearrange the order of the image sequence based on the Flatten type.
    if flatten_type is "row_major":
        pass 
    else:
        frames = transformation(frames, flatten_type)
        
    img = rearrange(frames, 
        "(n m) c h w -> c (n h) (m w)", 
        n=n, m=m)

    return img
\end{lstlisting}

The code above is the specific implementation of the Flatten method. 'Frames' represents a sequence of images that convey video information. Depending on the Flatten type, the sequence of 'frames' will be transformed, and then the transformed sequence of 'frames' will be flattened to achieve the conversion from an image sequence to a single image.

\subsection{Visualization}

\begin{figure*}
    \centering
    \includegraphics[width=0.9\linewidth]{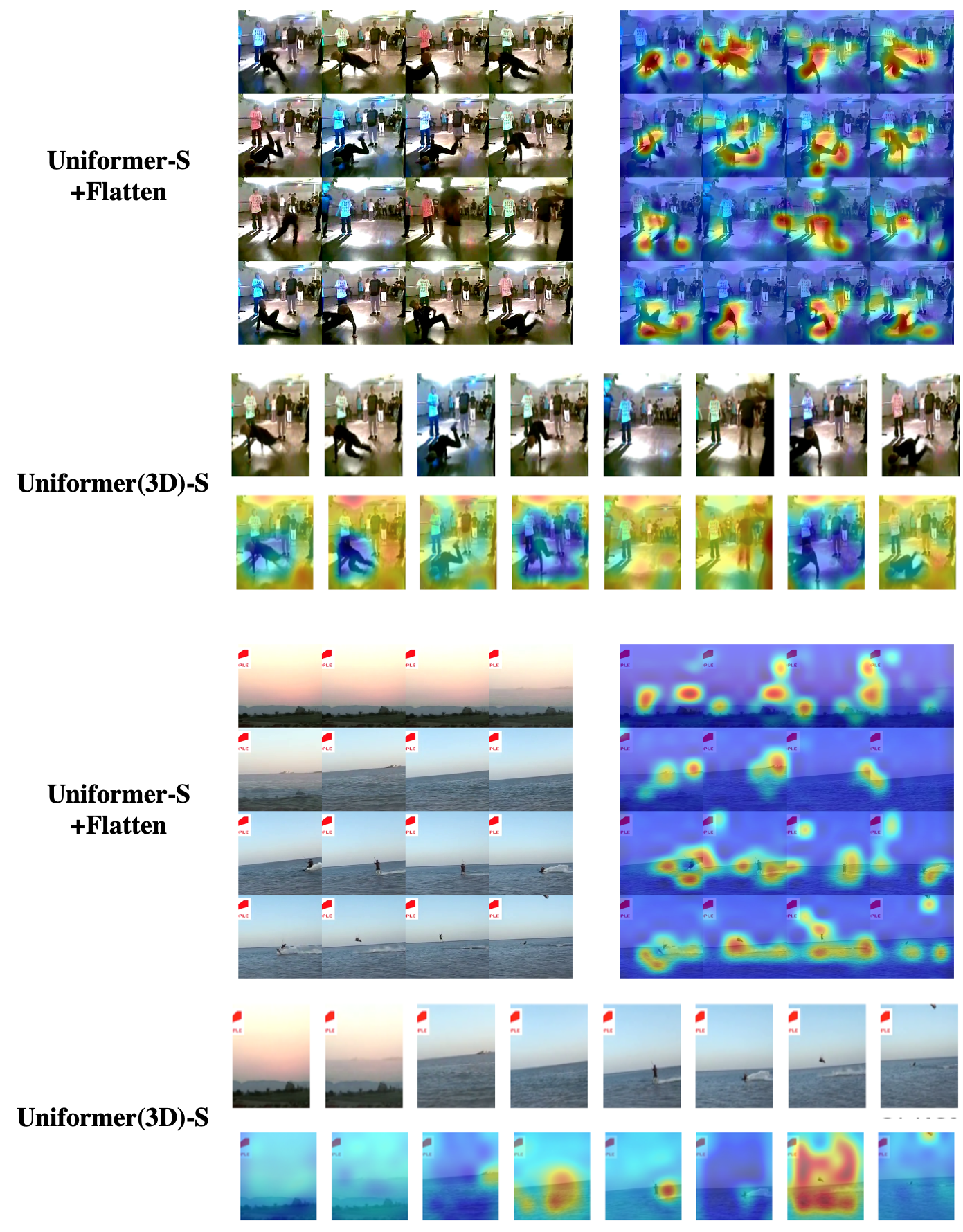}
    \caption{Heatmap visualization. The figure shows a comparison of the attention maps between the Uniformer-S+Flatten and Uniformer (3D)-S models."}
    \label{fig:heatmap_apd}
\end{figure*}

In Figure~\ref{fig:heatmap_apd}, we compare the attention maps of the Uniformer-S+Flatten and Uniformer (3D)-S models. It can be observed that Uniformer-S+Flatten has a greater focus on motion information and exhibits stronger attention. The visualization further demonstrates that the Flatten method aids the image understanding model in modeling temporal relationships through spatial positional information, thereby achieving a reduction in the boundary between image understanding and video understanding.

In Figure X, we show the changes in feature maps during the model's inference process. Compared to the original Uniformer (3D)-S, the Uniformer-S+Flatten model increasingly focuses on overall feature information as the network depth increases, thereby achieving global video temporal modeling.

\begin{figure*}[t]
    \centering
    \includegraphics[width=0.75\linewidth]{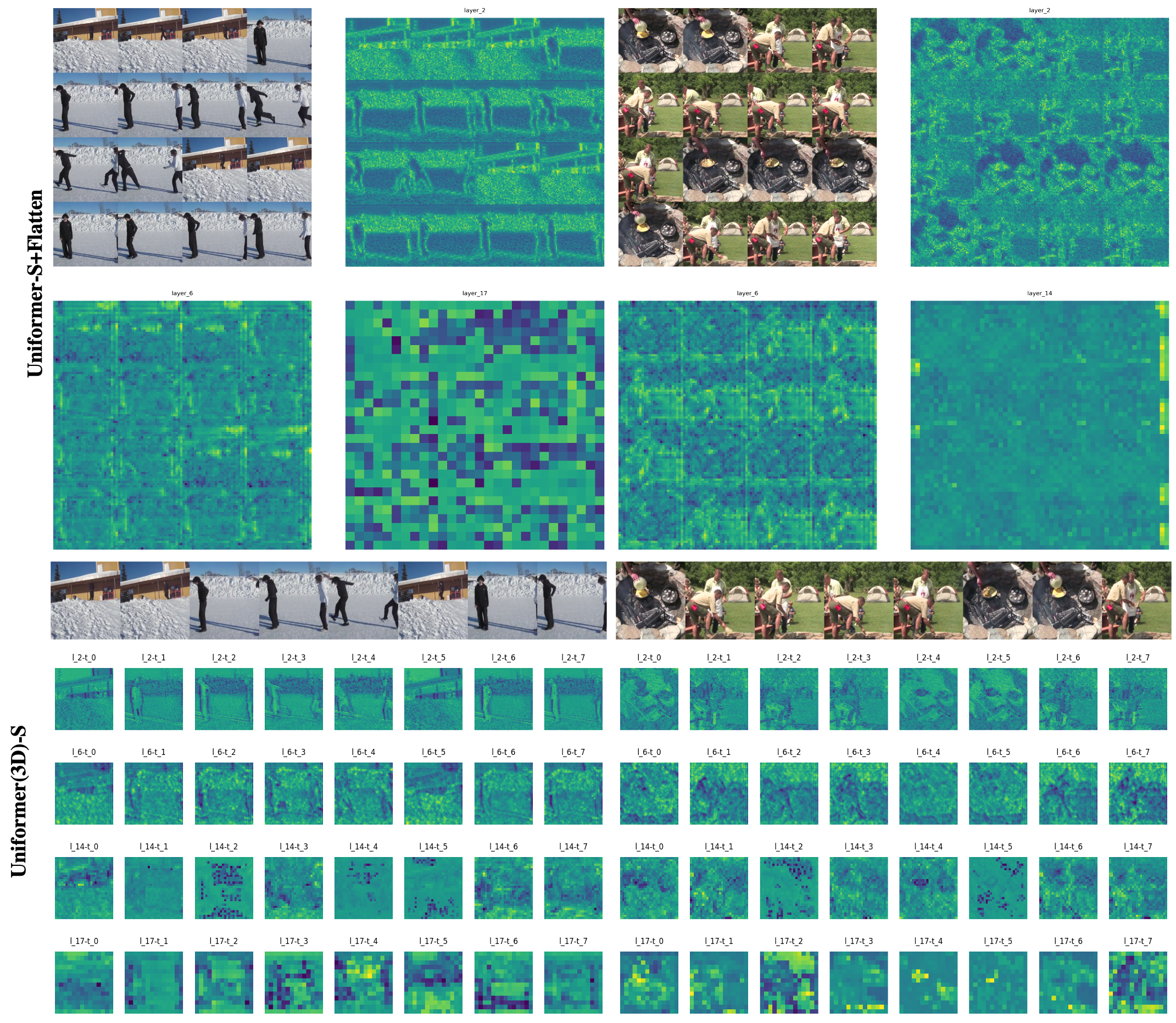   }
    \includegraphics[width=0.75\linewidth]{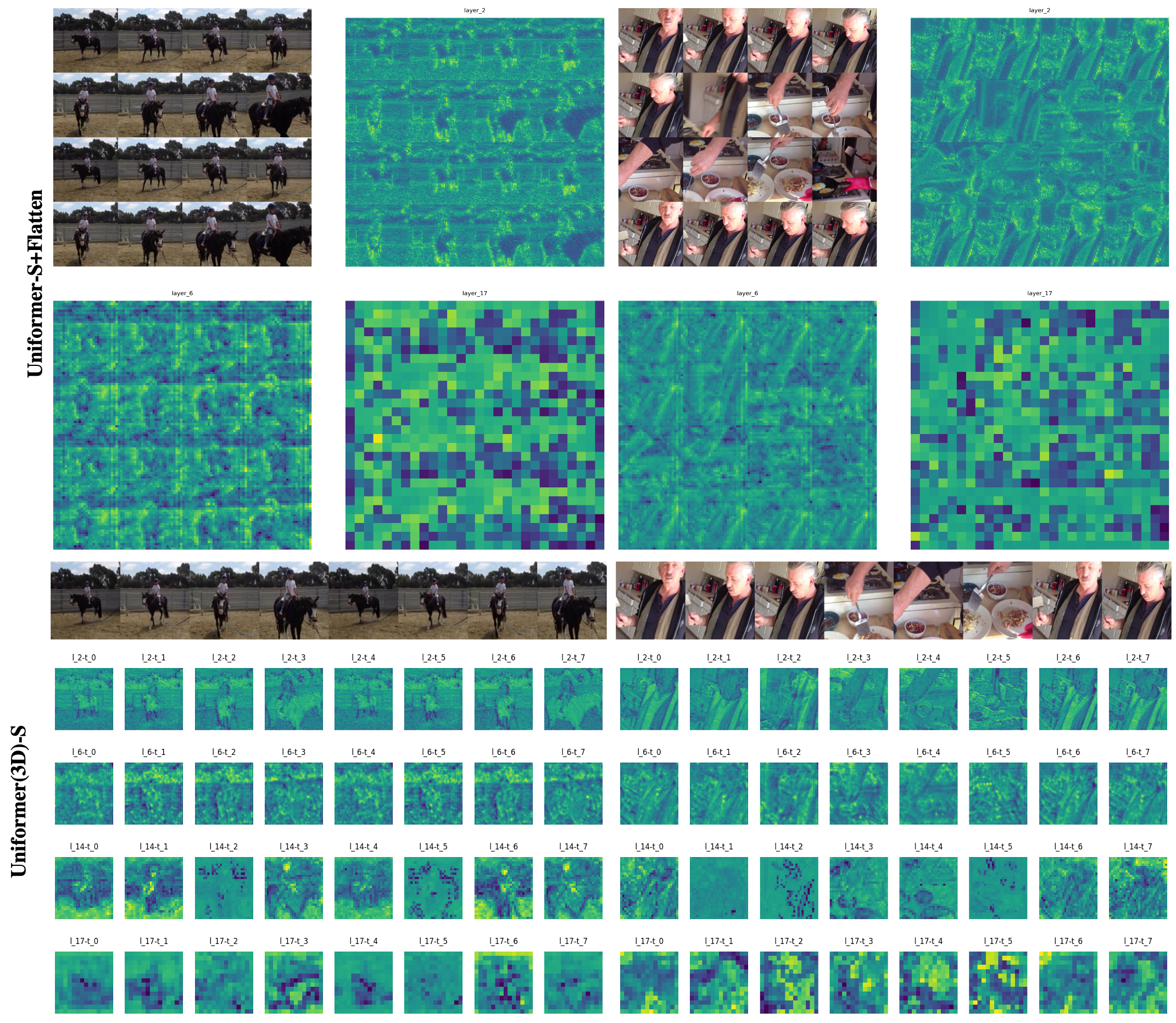   }
    \caption{Feature map visualization. The figure shows how the feature maps of the Uniformer-S+Flatten model change with network depth during the inference process. The deeper the network layer, the more global perspective the feature maps gain, providing a better overall understanding of the video’s temporal information."}
    \label{fig:fea_map}
\end{figure*}

\renewcommand{\arraystretch}{1.5}
\begin{table*}[ht]
    \centering
    \setlength{\tabcolsep}{1.17cm}
    \begin{tabular*}{\textwidth}{l c c c c}
    \hline
    \rowcolor{gray!20}
       \textbf{Model size} & \multicolumn{2}{c}{\textbf{Uniformer-XXS+Flatten}} & \multicolumn{2}{c}{\textbf{Uniformer-S+Flatten}} \\ \hline
    frames  & 9 & 16 & 9 & 16 \\
    \hline
        Optimiser &  \multicolumn{4}{c}{Adamw} \\
        Momentum & \multicolumn{4}{c}{0.9} \\
        Weight decay & \multicolumn{4}{c}{$5e^{-2}$} \\
        Batch size & 128 & 64 & 64 & 24 \\
        Learning rate schedule & \multicolumn{4}{c}{cosine with linear warmup} \\
        Linear warmup epochs & \multicolumn{4}{c}{10} \\
        Base learning rate & $5e^{-4}$ & $5e^{-4}$ & $1e^{-4}$ & $5e^{-5}$ \\
        Epochs & \multicolumn{4}{c}{110} \\ 
    \hline
    \end{tabular*}
    \caption{Hyperparameter information for different sizes of Uniformer+Flatten models on K400}
    \label{tab:1}
\end{table*}

\begin{table*}[p]
    \centering
    \setlength{\tabcolsep}{1.35cm}
    \begin{tabular*}{\textwidth}{l c c c }
    \hline
    \rowcolor{gray!20}
       \textbf{Datasets type} & \textbf{K400} & \textbf{SSV2} & \textbf{HMDB51} \\ \hline
    Model  & \multicolumn{3}{c}{Uniformer-S+Flatten} \\
    \hline
        Optimiser &  \multicolumn{3}{c}{Adamw} \\
        Momentum & \multicolumn{3}{c}{0.9} \\
        Weight decay & \multicolumn{3}{c}{$5e^{-2}$} \\
        Batch size & \multicolumn{3}{c}{24} \\
        Learning rate schedule & \multicolumn{3}{c}{cosine with linear warmup} \\
        Linear warmup epochs & 10 & ~~~~~~~~ 5 & 7 \\
        Base learning rate & $5e^{-5}$ & ~~~~~~~~ $1e^{-4}$ & $1e^{-4}$ \\
        Epochs & 110 & ~~~~~~~~ 25 & 70 \\ 
    \hline
    \end{tabular*}
    \caption{Training hyperparameters for Uniformer-S + Flatten model on different datasets}
    \label{tab:2}
\end{table*}

\begin{table*}[p]
    \centering
    \setlength{\tabcolsep}{0.72cm}
    \begin{tabular*}{\textwidth}{l c c c }
    \hline
    \rowcolor{gray!20}
       \textbf{Model type} & \textbf{Uniformer-S+Flatten} & \textbf{SwinV2-T+Flatten} & \textbf{ResNet101+Flatten} \\ \hline
    Dataset  & \multicolumn{3}{c}{K400} \\
    \hline
        Optimiser &  \multicolumn{3}{c}{Adamw} \\
        Momentum & \multicolumn{3}{c}{0.9} \\
        Weight decay & \multicolumn{3}{c}{$5e^{-2}$} \\
        Batch size & 24 & 24 & 64 \\
        Learning rate schedule & \multicolumn{3}{c}{cosine with linear warmup} \\
        Linear warmup epochs & \multicolumn{3}{c}{10} \\
        Base learning rate & $5e^{-5}$ & $1e^{-4}$ & $2e^{-4}$ \\
        Epochs & \multicolumn{3}{c}{110} \\ 
    \hline
    \end{tabular*}
    \caption{Training hyperparameters for different network models(Uniformer, SwinV2-T, ResNet) + Flatten operation on the K400 dataset}
    \label{tab:3}
\end{table*}

\clearpage
\bibliography{aaai25}

\begin{thebibliography}{32}
\providecommand{\natexlab}[1]{#1}

\bibitem[{Arnab et~al.(2021)Arnab, Dehghani, Heigold, Sun, Lu{\v{c}}i{\'c}, and Schmid}]{arnab2021vivit}
Arnab, A.; Dehghani, M.; Heigold, G.; Sun, C.; Lu{\v{c}}i{\'c}, M.; and Schmid, C. 2021.
\newblock Vivit: A video vision transformer.
\newblock In \emph{Proceedings of the IEEE/CVF international conference on computer vision}, 6836--6846.

\bibitem[{Bertasius, Wang, and Torresani(2021)}]{bertasius2021space}
Bertasius, G.; Wang, H.; and Torresani, L. 2021.
\newblock Is space-time attention all you need for video understanding?
\newblock In \emph{ICML}, volume~2, 4.

\bibitem[{Cao et~al.(2022)Cao, Wang, Chen, Jiang, Zhang, Tian, and Wang}]{cao2022swin}
Cao, H.; Wang, Y.; Chen, J.; Jiang, D.; Zhang, X.; Tian, Q.; and Wang, M. 2022.
\newblock Swin-unet: Unet-like pure transformer for medical image segmentation.
\newblock In \emph{European conference on computer vision}, 205--218. Springer.

\bibitem[{Carion et~al.(2020)Carion, Massa, Synnaeve, Usunier, Kirillov, and Zagoruyko}]{carion2020end}
Carion, N.; Massa, F.; Synnaeve, G.; Usunier, N.; Kirillov, A.; and Zagoruyko, S. 2020.
\newblock End-to-end object detection with transformers.
\newblock In \emph{European conference on computer vision}, 213--229. Springer.

\bibitem[{Carreira and Zisserman(2017)}]{carreira2017quo}
Carreira, J.; and Zisserman, A. 2017.
\newblock Quo vadis, action recognition? a new model and the kinetics dataset.
\newblock In \emph{proceedings of the IEEE Conference on Computer Vision and Pattern Recognition}, 6299--6308.

\bibitem[{Dai et~al.(2021)Dai, Liu, Le, and Tan}]{dai2021coatnet}
Dai, Z.; Liu, H.; Le, Q.~V.; and Tan, M. 2021.
\newblock Coatnet: Marrying convolution and attention for all data sizes.
\newblock \emph{Advances in neural information processing systems}, 34: 3965--3977.

\bibitem[{Dosovitskiy et~al.(2020)Dosovitskiy, Beyer, Kolesnikov, Weissenborn, Zhai, Unterthiner, Dehghani, Minderer, Heigold, Gelly et~al.}]{dosovitskiy2020vit}
Dosovitskiy, A.; Beyer, L.; Kolesnikov, A.; Weissenborn, D.; Zhai, X.; Unterthiner, T.; Dehghani, M.; Minderer, M.; Heigold, G.; Gelly, S.; et~al. 2020.
\newblock An image is worth 16x16 words: Transformers for image recognition at scale.
\newblock \emph{arXiv preprint arXiv:2010.11929}.

\bibitem[{Fan et~al.(2020)Fan, Li, Xiong, Lo, and Feichtenhofer}]{fan2020pyslowfast}
Fan, H.; Li, Y.; Xiong, B.; Lo, W.-Y.; and Feichtenhofer, C. 2020.
\newblock PySlowFast.
\newblock \url{https://github.com/facebookresearch/slowfast}.
\newblock Accessed: 2024-07-29.

\bibitem[{Feichtenhofer(2020)}]{feichtenhofer2020x3d}
Feichtenhofer, C. 2020.
\newblock X3d: Expanding architectures for efficient video recognition.
\newblock In \emph{Proceedings of the IEEE/CVF conference on computer vision and pattern recognition}, 203--213.

\bibitem[{Feichtenhofer et~al.(2019)Feichtenhofer, Fan, Malik, and He}]{feichtenhofer2019slowfast}
Feichtenhofer, C.; Fan, H.; Malik, J.; and He, K. 2019.
\newblock Slowfast networks for video recognition.
\newblock In \emph{Proceedings of the IEEE/CVF international conference on computer vision}, 6202--6211.

\bibitem[{Feichtenhofer, Pinz, and Zisserman(2016)}]{feichtenhofer2016convolutional}
Feichtenhofer, C.; Pinz, A.; and Zisserman, A. 2016.
\newblock Convolutional two-stream network fusion for video action recognition.
\newblock In \emph{Proceedings of the IEEE conference on computer vision and pattern recognition}, 1933--1941.

\bibitem[{Goyal et~al.(2017)Goyal, Ebrahimi~Kahou, Michalski, Materzynska, Westphal, Kim, Haenel, Fruend, Yianilos, Mueller-Freitag et~al.}]{goyal2017something}
Goyal, R.; Ebrahimi~Kahou, S.; Michalski, V.; Materzynska, J.; Westphal, S.; Kim, H.; Haenel, V.; Fruend, I.; Yianilos, P.; Mueller-Freitag, M.; et~al. 2017.
\newblock The" something something" video database for learning and evaluating visual common sense.
\newblock In \emph{Proceedings of the IEEE international conference on computer vision}, 5842--5850.

\bibitem[{He et~al.(2016)He, Zhang, Ren, and Sun}]{he2016deep}
He, K.; Zhang, X.; Ren, S.; and Sun, J. 2016.
\newblock Deep residual learning for image recognition.
\newblock In \emph{Proceedings of the IEEE conference on computer vision and pattern recognition}, 770--778.

\bibitem[{Kay et~al.(2017)Kay, Carreira, Simonyan, Zhang, Hillier, Vijayanarasimhan, Viola, Green, Back, Natsev et~al.}]{kay2017kinetics}
Kay, W.; Carreira, J.; Simonyan, K.; Zhang, B.; Hillier, C.; Vijayanarasimhan, S.; Viola, F.; Green, T.; Back, T.; Natsev, P.; et~al. 2017.
\newblock The kinetics human action video dataset.
\newblock \emph{arXiv preprint arXiv:1705.06950}.

\bibitem[{Kuehne et~al.(2011)Kuehne, Jhuang, Garrote, Poggio, and Serre}]{kuehne2011hmdb}
Kuehne, H.; Jhuang, H.; Garrote, E.; Poggio, T.; and Serre, T. 2011.
\newblock HMDB: a large video database for human motion recognition.
\newblock In \emph{2011 International conference on computer vision}, 2556--2563. IEEE.

\bibitem[{Li et~al.(2023)Li, Wang, Zhang, Gao, Song, Liu, Li, and Qiao}]{li2023uniformer}
Li, K.; Wang, Y.; Zhang, J.; Gao, P.; Song, G.; Liu, Y.; Li, H.; and Qiao, Y. 2023.
\newblock Uniformer: Unifying convolution and self-attention for visual recognition.
\newblock \emph{IEEE Transactions on Pattern Analysis and Machine Intelligence}, 45(10): 12581--12600.

\bibitem[{Liu et~al.(2020)Liu, Dou, Yu, and Heng}]{liu2020ms}
Liu, Q.; Dou, Q.; Yu, L.; and Heng, P.~A. 2020.
\newblock MS-Net: multi-site network for improving prostate segmentation with heterogeneous MRI data.
\newblock \emph{IEEE transactions on medical imaging}, 39(9): 2713--2724.

\bibitem[{Liu et~al.(2022{\natexlab{a}})Liu, Hu, Lin, Yao, Xie, Wei, Ning, Cao, Zhang, Dong et~al.}]{liu2022swin}
Liu, Z.; Hu, H.; Lin, Y.; Yao, Z.; Xie, Z.; Wei, Y.; Ning, J.; Cao, Y.; Zhang, Z.; Dong, L.; et~al. 2022{\natexlab{a}}.
\newblock Swin transformer v2: Scaling up capacity and resolution.
\newblock In \emph{Proceedings of the IEEE/CVF conference on computer vision and pattern recognition}, 12009--12019.

\bibitem[{Liu et~al.(2022{\natexlab{b}})Liu, Ning, Cao, Wei, Zhang, Lin, and Hu}]{liu2022video}
Liu, Z.; Ning, J.; Cao, Y.; Wei, Y.; Zhang, Z.; Lin, S.; and Hu, H. 2022{\natexlab{b}}.
\newblock Video swin transformer.
\newblock In \emph{Proceedings of the IEEE/CVF conference on computer vision and pattern recognition}, 3202--3211.

\bibitem[{Neimark et~al.(2021)Neimark, Bar, Zohar, and Asselmann}]{neimark2021video}
Neimark, D.; Bar, O.; Zohar, M.; and Asselmann, D. 2021.
\newblock Video transformer network.
\newblock In \emph{Proceedings of the IEEE/CVF international conference on computer vision}, 3163--3172.

\bibitem[{Selvaraju et~al.(2017)Selvaraju, Cogswell, Das, Vedantam, Parikh, and Batra}]{selvaraju2017grad}
Selvaraju, R.~R.; Cogswell, M.; Das, A.; Vedantam, R.; Parikh, D.; and Batra, D. 2017.
\newblock Grad-cam: Visual explanations from deep networks via gradient-based localization.
\newblock In \emph{Proceedings of the IEEE international conference on computer vision}, 618--626.

\bibitem[{Simonyan and Zisserman(2014)}]{Twostream}
Simonyan, K.; and Zisserman, A. 2014.
\newblock Two-stream convolutional networks for action recognition in videos.
\newblock \emph{Advances in neural information processing systems}, 27.

\bibitem[{Soomro, Zamir, and Shah(2012)}]{soomro2012ucf101}
Soomro, K.; Zamir, A.~R.; and Shah, M. 2012.
\newblock UCF101: A dataset of 101 human actions classes from videos in the wild.
\newblock \emph{arXiv preprint arXiv:1212.0402}.

\bibitem[{Tong et~al.(2022)Tong, Song, Wang, and Wang}]{tong2022videomae}
Tong, Z.; Song, Y.; Wang, J.; and Wang, L. 2022.
\newblock Videomae: Masked autoencoders are data-efficient learners for self-supervised video pre-training.
\newblock \emph{Advances in neural information processing systems}, 35: 10078--10093.

\bibitem[{Tran et~al.(2015)Tran, Bourdev, Fergus, Torresani, and Paluri}]{tran2015learning}
Tran, D.; Bourdev, L.; Fergus, R.; Torresani, L.; and Paluri, M. 2015.
\newblock Learning spatiotemporal features with 3d convolutional networks.
\newblock In \emph{Proceedings of the IEEE international conference on computer vision}, 4489--4497.

\bibitem[{Tran et~al.(2018)Tran, Wang, Torresani, Ray, LeCun, and Paluri}]{tran2018closer}
Tran, D.; Wang, H.; Torresani, L.; Ray, J.; LeCun, Y.; and Paluri, M. 2018.
\newblock A closer look at spatiotemporal convolutions for action recognition.
\newblock In \emph{Proceedings of the IEEE conference on Computer Vision and Pattern Recognition}, 6450--6459.

\bibitem[{Vaswani et~al.(2017)Vaswani, Shazeer, Parmar, Uszkoreit, Jones, Gomez, Kaiser, and Polosukhin}]{vaswani2017attention}
Vaswani, A.; Shazeer, N.; Parmar, N.; Uszkoreit, J.; Jones, L.; Gomez, A.~N.; Kaiser, {\L}.; and Polosukhin, I. 2017.
\newblock Attention is all you need.
\newblock \emph{Advances in neural information processing systems}, 30.

\bibitem[{Wang et~al.(2024{\natexlab{a}})Wang, Chen, Liu, Chen, Lin, Han, and Ding}]{wang2024yolov10}
Wang, A.; Chen, H.; Liu, L.; Chen, K.; Lin, Z.; Han, J.; and Ding, G. 2024{\natexlab{a}}.
\newblock Yolov10: Real-time end-to-end object detection.
\newblock \emph{arXiv preprint arXiv:2405.14458}.

\bibitem[{Wang et~al.(2021)Wang, Tong, Ji, and Wu}]{wang2021tdn}
Wang, L.; Tong, Z.; Ji, B.; and Wu, G. 2021.
\newblock Tdn: Temporal difference networks for efficient action recognition.
\newblock In \emph{Proceedings of the IEEE/CVF conference on computer vision and pattern recognition}, 1895--1904.

\bibitem[{Wang et~al.(2016)Wang, Xiong, Wang, Qiao, Lin, Tang, and Van~Gool}]{wang2016temporal}
Wang, L.; Xiong, Y.; Wang, Z.; Qiao, Y.; Lin, D.; Tang, X.; and Van~Gool, L. 2016.
\newblock Temporal segment networks: Towards good practices for deep action recognition.
\newblock In \emph{European conference on computer vision}, 20--36. Springer.

\bibitem[{Wang et~al.(2024{\natexlab{b}})Wang, Li, Li, Yu, He, Chen, Pei, Zheng, Xu, Wang et~al.}]{wang2024internvideo2}
Wang, Y.; Li, K.; Li, X.; Yu, J.; He, Y.; Chen, G.; Pei, B.; Zheng, R.; Xu, J.; Wang, Z.; et~al. 2024{\natexlab{b}}.
\newblock Internvideo2: Scaling video foundation models for multimodal video understanding.
\newblock \emph{arXiv preprint arXiv:2403.15377}.

\bibitem[{Zhang et~al.(2021)Zhang, Li, Liu, Shuai, Zhu, Brattoli, Chen, Marsic, and Tighe}]{zhang2021vidtr}
Zhang, Y.; Li, X.; Liu, C.; Shuai, B.; Zhu, Y.; Brattoli, B.; Chen, H.; Marsic, I.; and Tighe, J. 2021.
\newblock Vidtr: Video transformer without convolutions.
\newblock In \emph{Proceedings of the IEEE/CVF international conference on computer vision}, 13577--13587.

\end{thebibliography}

\end{document}